\documentclass[fullpage]{article}
\usepackage{color}
\usepackage{hyperref}
\definecolor{darkgreen}{rgb}{0,0.45,0}
\hypersetup{pdftex,colorlinks=true,urlcolor=blue,citecolor=darkgreen,linkcolor=darkgreen}
\usepackage{hypcap}
\pdfoutput=1
\usepackage{amsmath, amssymb}
\usepackage{mathtools}
\usepackage{xspace}
\usepackage{subcaption}
\usepackage{float}
\usepackage[capitalize]{cleveref}

\newcommand{\N}{\mathbb{N}}

\newcommand{\two}{\ensuremath{\mathbf{2}}\xspace}

\newcommand{\define}[1]{\textbf{\boldmath{#1}}}

\begin{document}

\title{Visualization tools for parameter selection in cluster analysis}

\author{Alexander Rolle and Luis Scoccola}

%
%


\maketitle

\abstract{
We propose an algorithm, HPREF (Hierarchical Partitioning by Repeated Features), 
that produces a hierarchical partition of a set of clusterings of a fixed dataset, 
such as sets of clusterings produced by running a clustering algorithm with a range of parameters. 
This gives geometric structure to such sets of clusterings, 
and can be used to visualize the set of results one obtains by running a clustering algorithm 
with a range of parameters.
}

\section{Introduction}

Often, a clustering algorithm, rather than producing a single clustering of a dataset,
produces a set of clusterings. For example, one gets a set of clusterings by running a clustering algorithm
with a range of parameters. 
The starting point of this paper is the observation that sets of clusterings ought to have geometric structure.
Indeed, various metrics have been proposed for the set of all clusterings of a fixed dataset \cite{meila, vanDongen, mirkin}.

In this paper, we define a metric on any set $S$ of clusterings of a fixed dataset that is particularly convenient for visualization. 
The metric is induced by a hierarchical partition of $S$, which is defined as follows. 
Any pair of data points $(x,y)$ can be used to partition $S$ into two classes: a class containing 
those clusterings that cluster together $x$ and $y$, and a class containing those that do not.
Say that two pairs of data points are equivalent if they define the same partition of $S$.
A large equivalence class defines a partition of $S$ that is witnessed by many pairs of data points. 
We produce a hierarchical partition of $S$ by successively partitioning $S$ according to the 
largest equivalence classes. 
Using pairs of data points to discriminate between different clusterings has a long history, 
particularly in the many variations on the so-called Rand index  \cite{rand,fowlkes,wallace,hubert}. 
The voting-style method of this paper is a practical, scalable 
way to adapt these ideas for the detection and visualization  
of the large-scale features of a set of clusterings.

In \cref{algorithm} we describe this procedure in detail, 
and in \cref{examples} demonstrate how our algorithm can be used as a visualization tool.
An implementation of HPREF is available at \cite{code}.

\section{Hierarchical Partitioning by repeated features} \label{algorithm}

By a \define{clustering} of a set $X$, we mean a set of disjoint subsets of $X$. 
Points of $X$ that do not belong to any subset in a clustering of $X$ 
are called \define{noise points}. 

We begin by recalling a well-known way to encode clusterings as binary vectors. 
Given a set $S$ of clusterings of $X$, there is an embedding
\[
    M : S \to \two^{P} \, ,
\]
where $\two = \{0,1\}$, $P$ is the set of unordered pairs of points of $X$ (with repetition),
and $\two^P$ is the set of binary vectors indexed by $P$.
For $C \in S$, let $M(C)_{(x,y)} = 0$ if $C$ clusters $x$ and $y$ together, and $M(C)_{(x,y)} = 1$ otherwise.
We allow pairs with repetition to distinguish noise points from one-point clusters: 
if $x \in X$ is a noise point of $C$, then $M(C)_{(x,x)} = 1$, but if $\{x\}$ is a cluster of $C$, then $M(C)_{(x,x)} = 0$. 
Write $s = |S|$ and $p = |P|$.
We'll think of $M(S)$ as an $s \times p$ matrix, so that a row of $M(S)$ is $M(C)$ for some $C \in S$.
Each column of $M(S)$ corresponds to a pair of points of $X$.

It is not usually practical to consider all pairs of data points when constructing the matrix $M(S)$. 
Instead, one can first sample pairs from the dataset, 
then construct $M(S)$ with columns corresponding only to the sampled pairs. 
Because HPREF uses columns of $M(S)$ that occur often, 
the method is robust with respect to this sampling; 
see \cref{example2} for experimental results.

\paragraph{Hierarchical partitioning.}
\label{a-hierarchical-approach}

A \define{hierarchical partition} of a set $S$ is a tree,
where each node is associated to a subset of $S$, such that the root is associated to $S$,
and the set associated to any node is a subset of the set associated to its parent.

A non-constant binary vector is a vector that contains both zeroes and ones.
Each non-constant column $c$ of a binary matrix $M$ partitions the set of rows of $M$ into two classes: the class of rows
with a zero in column $c$, and the class of rows with a one in column $c$.
If $R$ is the set of rows of $M$,
let us denote these two classes of rows by $R^c_0$ and $R^c_1$ respectively.
Let $f$ be a scoring function that assigns a numeric score to any set of clusterings of a fixed dataset.
We discuss the choice of scoring function below.

As input, the algorithm takes a set $S$ of clusterings of a fixed dataset, and $\max_l \in \N$.
The output is a hierarchical partition of $S$.
\begin{enumerate}
    \item[1.] Initialize a binary tree $T$ with just one node, and associate the set $S$ to it;
    \item[2.] While the number of leaves of $T$ is less than $\max_l$:
        \begin{enumerate}

    \item[3.] For each leaf $L$ of $T$, let $R_L \subseteq S$ be its associated set, and define the score
        of $L$ to be $f(R_L)$;
    \item[4.] Let $K$ be the leaf with the highest score, and let $R \subseteq S$ be its associated set;
    \item[5.] Let $c$ be the most repeated non-constant column of the matrix $M(R)$, and partition $R$ into the classes $R^c_0$ and $R^c_1$;
    \item[6.] Add two children to $K$, one with associated set $R^c_0$, and the other with associated set $R^c_1$.
        \end{enumerate}
    \item[7.] Return $T$.
\end{enumerate}

\paragraph{Dendrograms.}
Hierarchical partitions are especially useful when they can can be represented as dendrograms.
By \define{dendrogram} we mean a hierarchical partition of a set, where each node has a weight
such that the weight of any node is smaller than the weight of its parent.
The weights allow us to visualize the dendrogram in two dimensions.

The output of HPREF can be represented by a dendrogram:
let the weight of a node $n$ be given by the score of $n$ plus the sum
of the scores of all the nodes that were added to the tree after $n$.

Moreover, by a well-known construction (see, e.g., \cite{carlsson-memoli}), 
this dendrogram defines a metric on $S$.

\paragraph{Scoring functions.}

The goal of the scoring function is to quantify how much a set of clusterings deserves to be partitioned.

Say we are given a set of clusterings $S$, and form the binary matrix $M(S)$.
Let $m\in \N$ be the multiplicity of the most repeated non-constant column of $M(S)$, 
and let $c$ be the number of non-constant columns of $M(S)$.
A large value of $m$ indicates a partition of $S$ that is witnessed by many pairs of data points, 
and a large value of $c$ indicates heterogeneity in $S$. 
HPREF uses the scoring function
\[
    f(S) = c + m\,.
\]

\paragraph{Complexity.}
Let $n$ be the number of pairs of points of $X$ that we choose to sample,
and let $s = |S|$. Assume that $s \leq n$.
Using dictionaries implemented as tries, the time complexity of HPREF is in $O(n\times s \times \max_l)$.
The same analysis shows that this is also the space complexity.

\section{Examples} \label{examples}

In this section we present two examples. In the first, we generate a set of clusterings of Fisher's Iris dataset by running DBSCAN with a range of parameters, and show how one can visualize this set of clusterings using HPREF. Our algorithm allows one to easily identify the parameters for which DBSCAN separates the three species of Iris in the dataset.

In the second example, we generate a set of clusterings of a large dataset used for 
The Third International Knowledge Discovery and Data Mining Tools Competition, 
by running $k$-means with different initializations. 
This example shows that, even using a very small sample of pairs of data points,
our algorithm produces meaningful results.

For the examples, we use the scikit-learn implementations of DBSCAN and $k$-means.

\subsection{Clustering the Iris dataset with DBSCAN}

In a survey paper on density-based clustering by Kriegel, Kr\"{o}ger, Sander, and Zimek, 
the authors write that density-based clustering algorithms are ``particularly suitable'' 
for certain applications coming from biology \cite[p232]{KKSZ}; 
an example they give is Fisher's Iris dataset \cite{iris}, which illustrates the 
``typical properties of natural (biological) clusters'' \cite[p233]{KKSZ}. 
The Iris dataset  records the petal and sepal width and length of 150 Iris flowers. 
There are $50$ observations of each of the species Setosa, Versicolor, and Virginica; 
the observations of Setosa are linearly separable from the observations of Versicolor and Virginica, 
but the latter two are not linearly separable from each other.

The clustering algorithm DBSCAN may be the best known density-based clustering algorithm, and is a main topic of \cite{KKSZ}. 
It takes two parameters: a distance scale $\epsilon > 0$, and a density threshold $k \in \N$. 
We use HPREF to study the output of DBSCAN on the Iris dataset, as the parameters vary.

Let $T$ be the set of clusterings of the Iris dataset obtained by running DBSCAN with 
$(\epsilon,k) \in \{0.05 \cdot i \, | \, 1 \leq i \leq 20\} \times \{1, 2, \dots, 10\}$. 

\paragraph{HPREF.}

\begin{figure}[htbp]
    \centering
    \includegraphics[width=0.25\textwidth]{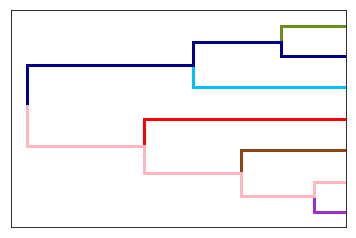}
    \caption{Dendrogram obtained by running HPREF on $T$ with $\max_l = 7$,
    colored according to the partitions of \cref{fig:iris-dendrogram-cuts}.}
    \label{fig:iris-dendrogram}
\end{figure}

The hierarchical clustering of $T$ produced by HPREF with $\max_l = 7$ 
is shown in \cref{fig:iris-dendrogram}.
The most repeated column of $M(T)$ appears $1170$ times. Recall that, at each node of the hierarchy, we are considering a matrix obtained from $M(T)$ by selecting a class of rows; the most repeated column of these matrices, in the order they appear in the hierarchy, appears $1170$, $349$, $240$, $273$, $226$, $251$ times.

\begin{figure}[htbp]
    \centering
    \includegraphics[width=0.142\textwidth]{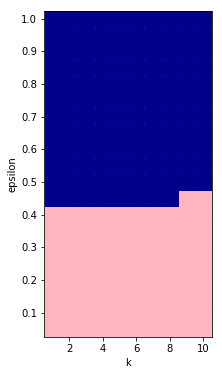}
    \includegraphics[width=0.12\textwidth]{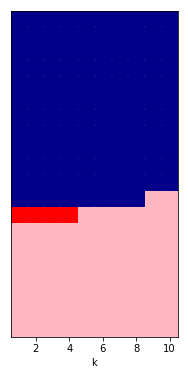}
    \includegraphics[width=0.12\textwidth]{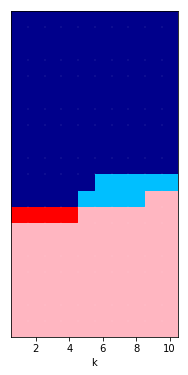}
    \includegraphics[width=0.12\textwidth]{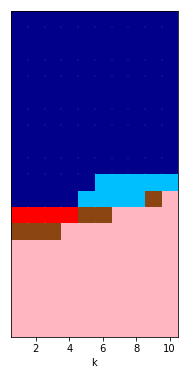}
    \includegraphics[width=0.12\textwidth]{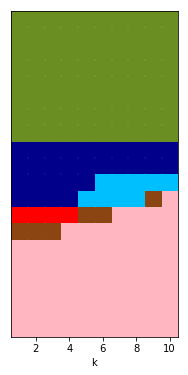}
    \includegraphics[width=0.12\textwidth]{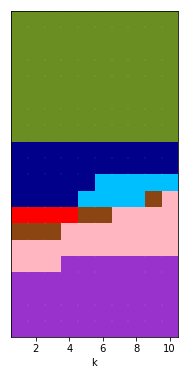}
    \caption{Partitions of $T$ corresponding to the $6$ possible vertical cuts of the dendrogram of \cref{fig:iris-dendrogram}.}
    \label{fig:iris-dendrogram-cuts}
\end{figure}

\begin{figure}[htbp]
    \centering
    \includegraphics[width=0.35\textwidth]{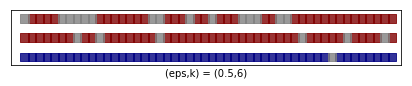}
    \includegraphics[width=0.35\textwidth]{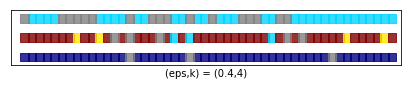}\\
    \includegraphics[width=0.35\textwidth]{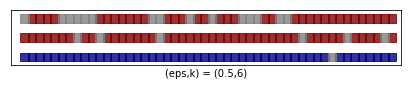}
    \includegraphics[width=0.35\textwidth]{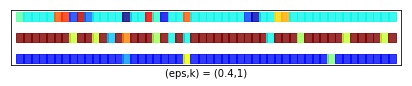}\\
    \includegraphics[width=0.35\textwidth]{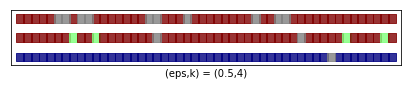}
    \includegraphics[width=0.35\textwidth]{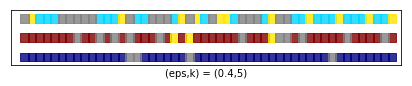}\\
    \includegraphics[width=0.35\textwidth]{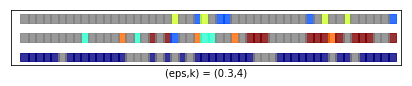}
    \includegraphics[width=0.35\textwidth]{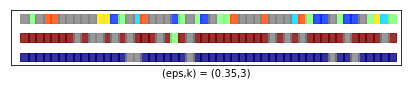}\\
    \includegraphics[width=0.35\textwidth]{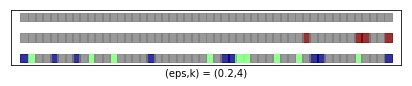}
    \caption{The Iris dataset, clustered by DBSCAN with several choices of parameters. 
    In each clustering, the bottom row represents the observations of Setosa, 
    the middle row Versicolor, 
    and the top row Virginica.}
    \label{fig:iris-clusterings}
\end{figure}

The red class of \cref{fig:iris-dendrogram-cuts}, which consists of the clusterings obtained with $(\epsilon,k) = (0.4,1), \dots, (0.4,4)$, 
contains the best solutions to the clustering problem: the clusterings in this class do a reasonable job of separating the three species of Iris.  
The brown class of \cref{fig:iris-dendrogram-cuts}, consisting of the clusterings obtained with 
\[
	(\epsilon, k) = \{(0.35,1),(0.35,2),(0.35,3),(0.4,5),(0.4,6),(0.45,9)\} \, ,
\]
also contains interesting results, 
but these clusterings are further from the ``correct'' clustering of the dataset, 
as they separate the observations of Iris Virginica into multiple clusters.
See \cref{fig:iris-clusterings} for representatives of the different classes.

Since the Iris dataset is labeled, we can compare the clusterings produced by DBSCAN with the labels,
using one of the standard distances between clusterings. 
In \cref{table:randindex} we compute the average, maximum, minimum, and standard deviation of the
adjusted Rand index (\cite{hubert}) between the clusterings of each of the classes of
the second partition of \cref{fig:iris-dendrogram-cuts} and the labels.
We regard noise points as one-point clusters when computing the adjusted Rand index.
We see that, according to the adjusted Rand index, the red class coincides with the best four clusterings of $T$.

\begin{center}
\begin{table}[t]%
\centering
\begin{tabular}{ |c|c|c|c|c|c| }
	\hline
    {\small \# clusterings} & {\small adj. Rand} & {\small adj. Rand} & {\small adj. Rand} & {\small adj. Rand} \\
	{\small in class} & {\small mean} & {\small min.} & {\small max.} & {\small std. deviation} \\
		\hline
    {\small $4$} & {\small $0.699$} & {\small $0.684$} 
    & {\small $0.706$} & {\small $0.008$}  \\
	\hline
    {\small $118$} & {\small $0.549$} & {\small $0.465$} 
    & {\small $0.568$} & {\small $0.022$} \\
	\hline
    {\small $78$} & {\small $0.168$} & {\small $0$} & {\small $0.589$} & {\small $0.194$} \\ 
    \hline
\end{tabular}

\caption{\label{table:randindex}Adjusted Rand index between the clusterings of $T$ and the true labels of the
    Iris dataset, partitioned by the second partition of \cref{fig:iris-dendrogram-cuts}. From top to bottom,
    red, blue, and pink classes.}

\end{table}
\end{center}

\begin{figure}[htbp]
    \centering
    \includegraphics[width=0.25\textwidth]{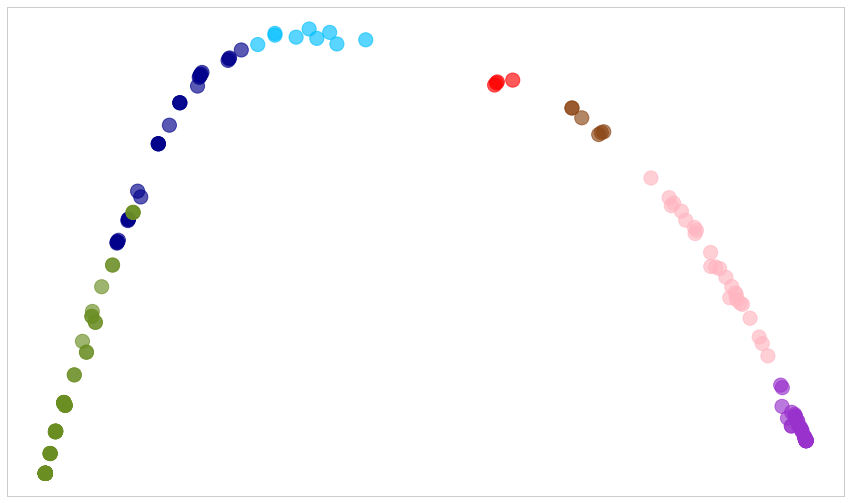}
    \caption{PCA with $2$ components applied to $M(T)$, colored by the last partition of \cref{fig:iris-dendrogram-cuts}.}
    \label{fig:irispca}
\end{figure}

\paragraph{Alternative visualization of the space of clusterings.}
We apply PCA to the rows of $M(T)$. We keep the first $2$ components, which account for approximately $78\%$ of the variance.
We plot the first $2$ components of the rows in \cref{fig:irispca}, colored according to the
last partition of \cref{fig:iris-dendrogram-cuts}.
In this example, HPREF captures much of the geometric structure that we see in the visualization produced by PCA:
the partitions of $T$ produced by HPREF correspond well to the clustering structure we see in the visualization,
and the order in which these distinctions appear in the hierarchy reflect the extent to which these distinctions 
are obvious in the visualization.

\subsection{Choosing initial centers for $k$-means}
\label{example2}

Given a finite set of points in euclidean space, the $k$-means problem is to choose $k$ centers that minimize $\phi$,
the sum of the squared distance between each point and its closest center.
A commonly used algorithm to find approximate solutions to the $k$-means problem is due to Lloyd \cite{lloyd}.
The algorithm begins by choosing $k$ centers at random from the dataset.
It then assigns each data point to its closest center, and recomputes each center as the center of mass of the
points assigned to it. This step is repeated until the process stabilizes, to obtain $k$ centers $x_1, \ldots, x_k$.
This produces a clustering with $k$ clusters, for which a point $x$ belongs to the $i^{th}$ cluster if
the closest center to $x$ is $x_i$.

Of course, the outcome of Lloyd's algorithm depends on the choice of the initial centers.
A common approach is to choose these initial centers uniformly at random from the dataset.
In \cite{arthur-vassilvitskii}, Arthur and Vassilvitskii propose a more sophisticated approach:
choosing the initial centers at random from the dataset, but weighing data points according to their
squared distance from the closest center already chosen.

Following \cite{arthur-vassilvitskii}, we'll refer to Lloyd's algorithm, with initial centers chosen uniformly at random from the dataset, 
as \texttt{k-means}, and we'll refer to Lloyd's algorithm, with initial centers chosen according to the method of \cite{arthur-vassilvitskii}, 
as \texttt{k-means++}.

In \cite{arthur-vassilvitskii}, Arthur and Vassilvitskii compare the performance of 
\texttt{k-means} and \texttt{k-means++} on four datasets, including the KDD Cup 1999 dataset from the 
University of California--Irvine Machine Learning Repository \cite{intrusion}.

The KDD Cup 1999 dataset simulates features available to an intrusion detection system, 
and was the dataset used for The Third International Knowledge Discovery and Data Mining Tools Competition. 
We used the full dataset available at the UCI Machine Learning Repository, 
which consists of $4,898,431$ points. We kept the $34$ continuous features, ignoring 
the $8$ categorical features. 
Following \cite{arthur-vassilvitskii}, we consider a set $U$ of 40 clusterings of the KDD Cup 1999 dataset,
with 20 produced by \texttt{k-means} and 20 produced by \texttt{k-means++}, 
both with $k = 25$.
Information about the associated values of $\phi$ is in \cref{table:intrusion_statistics}.

\begin{figure}[htbp]
    \centering
    \includegraphics[width=0.25\textwidth]{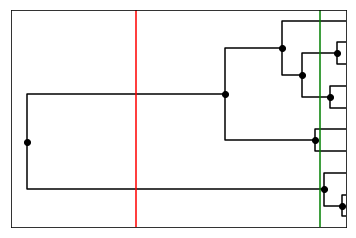}
    \caption{Dendrogram obtained by running HPREF on $U$ with $\max_l = 10$.}
    \label{fig:intrusion-dendrogram}
\end{figure}

\paragraph{HPREF.} We run HPREF on $U$ with $\max_l = 10$ and a sample of $20,000$ pairs of data points. 
The resulting hierarchy is displayed in \cref{fig:intrusion-dendrogram}.
The most repeated column of $M(U)$ appears $8,340$ times, and exactly separates the output of \texttt{k-means} and \texttt{k-means++}. 
I.e., the partition of $U$ corresponding to the red cut of \cref{fig:intrusion-dendrogram} has one class containing the clusterings produced by \texttt{k-means}, and another class containing the clusterings produced by \texttt{k-means++}.

To get a finer partition of $U$, we consider the partition corresponding to the green cut of \cref{fig:intrusion-dendrogram}. This is the finest partition produced by HPREF that does not divide the output of \texttt{k-means++} into multiple classes. 
The elements of $U$ produced by \texttt{k-means} are partitioned into five classes, three of which are singletons. 
The results are displayed in \cref{table:partitioning_intrusion}.

\begin{center}
\begin{table}[t]%
\centering
\begin{tabular}{ |l|c|c|c|c|c| }
	\hline
	{\small clusterings} & {\small number of} & {\small $\phi$} & {\small $\phi$} & {\small $\phi$} & {\small $\phi$} \\
	& {\small clusterings} & {\small mean} & {\small min.} & {\small max.} & {\small std. deviation} \\
	\hline
    {\small $U$} & {\small 40} & {\small $3.537 \cdot 10^{14}$} & {\small $7.827 \cdot 10^{13}$} 
    & {\small $7.339 \cdot 10^{14}$} & {\small $2.915 \cdot 10^{14}$} \\
	\hline
    {\small \texttt{k-means}} & {\small 20} & {\small $6.279 \cdot 10^{14}$} & {\small $2.286 \cdot 10^{14}$} 
    & {\small $7.339 \cdot 10^{14}$} & {\small $1.397 \cdot 10^{14}$} \\
	\hline
    {\small \texttt{k-means++}} & {\small 20} & {\small $7.946 \cdot 10^{13}$} & {\small $7.827 \cdot 10^{13}$} 
    & {\small $8.335 \cdot 10^{13}$} & {\small $1.935 \cdot 10^{12}$} \\
	\hline
\end{tabular}

\caption{\label{table:intrusion_statistics} Experimental results of \texttt{k-means}
          and \texttt{k-means++} on the KDD Cup 1999 dataset, with $k=25$.}
\end{table}
\end{center}

\begin{center}
\begin{table}[t]%
\centering
\begin{tabular}{ |c|c|c|c|c|c| }
	\hline
    {\small \# clusterings} & {\small $\phi$} & {\small $\phi$} & {\small $\phi$} & {\small $\phi$} \\
	{\small in class} & {\small mean} & {\small min.} & {\small max.} & {\small std. deviation} \\
		\hline
    {\small 4} & {\small $6.227 \cdot 10^{14}$} & {\small $6.191 \cdot 10^{14}$} 
    & {\small $6.336 \cdot 10^{14}$} & {\small $6.242 \cdot 10^{12}$} \\
	\hline
    {\small 13} & {\small $6.839 \cdot 10^{14}$} & {\small $6.114 \cdot 10^{14}$} 
    & {\small $7.339 \cdot 10^{14}$} & {\small $4.140 \cdot 10^{13}$}  \\
	\hline
    {\small 1} & {\small $2.288 \cdot 10^{14}$} & {\small $2.288 \cdot 10^{14}$} & {\small $2.288 \cdot 10^{14}$} & -- \\
	\hline
    {\small 1} & {\small $2.286 \cdot 10^{14}$} & {\small $2.286 \cdot 10^{14}$} & {\small $2.286 \cdot 10^{14}$} & -- \\
	\hline
    {\small 1} & {\small $7.188 \cdot 10^{14}$} & {\small $7.188 \cdot 10^{14}$} & {\small $7.188 \cdot 10^{14}$} & -- \\
	\hline
     {\small 20} & {\small $7.946 \cdot 10^{13}$} & {\small $7.827 \cdot 10^{13}$} 
    & {\small $8.335 \cdot 10^{13}$} & {\small $1.935 \cdot 10^{12}$} \\
\hline

\end{tabular}
\caption{\label{table:partitioning_intrusion} The partition of the elements of $U$ corresponding 
    to the green cut of \cref{fig:intrusion-dendrogram}. The last class contains exactly the \texttt{k-means++} runs.}
\end{table}
\end{center}

We see that the partition corresponds well to the $\phi$ values of the clusterings.
In particular, while the standard deviation of the $\phi$ values of the \texttt{k-means} runs is on the order
of $10^{14}$, the standard deviation of the $\phi$ values in each class is on the order of $10^{13}$ or less.

\paragraph{Sampling and performance.}
To test the reliability of this result, we run HPREF on $100$ samples of $20,000$ pairs of data points, 
each time with $\max_l = 6$.
In every case, the partition obtained by taking the leaves of the hierarchy is exactly the result displayed in \cref{table:partitioning_intrusion}. 
We also run HPREF on $100$ samples of $5,000$ pairs of data points, and obtain the result of 
\cref{table:partitioning_intrusion} $99$ times.

On a laptop with a 2.20 GHz Intel Core i7 and 16GB of RAM, 
using the scikit-learn implementation of \texttt{k-means} and \texttt{k-means++}, 
it took $4$ hours and $32$ minutes to 
generate the set $U$ of $40$ clusterings of the KDD Cup 1999 dataset.
Running HPREF on $100$ samples of $20,000$ pairs of data points took $12$ minutes. 
Running HPREF on $100$ samples of $5,000$ pairs of data points took $3$ minutes.

\section*{Acknowledgments}

We would like to thank Dan Christensen, Camila de Souza, and Rick Jardine
for their helpful comments and suggestions.

\bibliography{RolleScoccolaRef}
\bibliographystyle{plain}

\end{document}